# Futuristic Classification with Dynamic Reference Frame Strategy

[a*] Kumarjit Pathak, [b*]Jitin Kapila, [c]Aasheesh Barvey

**Abstract** Classification is one of the widely used analytical techniques in data science domain across different business to associate a pattern which contribute to the occurrence of certain event which is predicted with some likelihood. This Paper address a lacuna of creating some time window before the prediction actually happen to enable organizations some space to act on the prediction.
There are some really good state of the art machine learning techniques to optimally identify the possible churners in either customer base or employee base, similarly for fault prediction too if the prediction does not come with some buffer time to act on the fault it is very difficult to provide a seamless experience to the user. New concept of reference frame creation is introduced to solve this problem in this paper

*Index Terms*— **Prediction, Classification, Churn prediction, Reference Frame, Advance modelling, Early warning system, Prediction window, Deep learning**

## I. INTRODUCTION

In current market situation proactive engagement is highly important to either retain customers who are likely to churn. Similarly prediction of fault by device is highly important however for both the modelling if organization does not get some window where it can act to either engage with customer to stop churn or engage to rectify the fault then the prediction is not of much use as th damage is already done and there is not lead time to action on anything to manage the customer loss or the reliability loss of user experience.

In response to this problem we propose a methodology to detect the event occurrence with some agreed lead time.

## II. LITERATURE REVIEW AND RELATED WORK

Current research focus is mainly to achieve a better classification model to accurately detect the even be it churn or fault occurrence.

Clifton Phua et al[2] has used several machine learning model LADTree, Decision Stump, RepTree, J48, NaiveBayes, TreeLMT, RandomForest, Bagging+BF Tree, Bagging+LADTree, SimpleCart, classification Via Regression to find out potential churners and predict possible winback as well in near future.

V. Umayaparvathi et al[3] emphasize on the data and variable engineering and usage of multiple techniques to detect churn/customer complaint/network fault propensity etc.

Jingjiao Zhang et al[1] has worked on this direction to create early detection system to study the effectiveness to find the churners as early as possible with the accuracy being high enough, which is defined as Early Churn Prediction. The predictive performance of the proposed model, which takes time series attributes and influence of churning contacts in social network into consideration, is investigated. We evaluate the method using a 12-month-long dataset collected by one of the largest operators in China.

This work is based on time series impact analysis on churn however taking time series as a prediction tends to create auto correlation impact to the classification.

Amoo A O et al[4] has used Adaptive Neuro Fuzzy Inference System – based prediction model for customer churn in telecommunication industry was emulated. Exhaustive search algorithm was employed for feature selection of the most significant variables influencing churn tendency. This was with a view to identifying the key performance indicators for churn tendency. Fuzzy rules were set up to represent the antecedents with their corresponding consequents.

Imran Khan et al[5] has used Genetic Programming (GP) to evolve a suitable classifier by using the customer based features. Genetic Programming (GP) is population based heuristic used to solve complex multimodal optimization problems. It is an evolutionary approach use the Darwinian

Authors:
[a*] Mr. Kumarjit Pathak, Data Scientist, Working @ Harman Connected Services India Pvt Bengaluru, Karnataka- 560066 (e-mail: Kumarjit.pathak@ outlook.com).
[b*] Mr. Jitin Kapila, Data Scientist, Working @ Zetaglobal, Bengaluru, Karnataka- 560066 (e-mail: Jitin.kapila@ outlook.com).

[c] Mr. Aasheesh Barvey, Data Scientist Working @ Harman Connected Services India Pvt Ltd., Bengaluru, Karnataka- 560066 (e-mail: ashbarvey@ gmail.com).

\* Major and equal contribution

principle of natural selection (survival of the fittest) analogs with various naturally occurring operations, including crossover (sexual recombination), mutation (to randomly perturbed or change the respective gene value) and gene duplication.

Pradeep B et al[7] has explained about different supervised and unsupervised technique of machine learning which helped to detect churn for logistic industry perspective.

Shin-Yuan Hung et al[11] has worked on decision tree architecture to predict the propensity to churn.

After going through different research paper published and material available on the internet one unexplored area of research appears. This is majorly on how algorithm can ensure that classification models can predict the propensity of churn or fault or similar classification with some lead time for the organization to act. Method depicted in this paper aims to unveil the strategy to a solution.

### III. SOLUTION ARCHITECTURE

Statistical model building for any given type of problem includes 'data', 'learning parameter'. However here in this paper we are introducing concept of "reference frame", which also an inherent feature which the algorithm learns.

This concept is described in Physics by Thomas DeMichele et al[18] as "Frame of reference: A point of view, context, or set of coordinates with which we can orient ourselves. This can be a person's physical viewpoint or can be space-time coordinates we calibrate an instrument to. It is a defined vantage-point that we can observe/measure from".

The regular data we see in the either in regular classification modelling is mostly cross sectional data. Where we are modelling to classify an observation with a propensity measure. During the machine learning modelling exercise, the algorithm learns to identify due to what range of values in the independent variable set the outcome events observed. Thus defining a boundary of the values in the set of the independent variables, which if occurred can be associated to a class.

Like if the algorithm sees in Fraud Modelling exercise, non-payment of monthly loan amount it would comfortably classify the observation as fraud.

In case of fault classification model if the network device is having a certain threshold value to packet drops then the device is classified as faulty.

In the example of churn analysis if the customer purchase has dropped in last few months drastically. The algorithm might assign high propensity for the mentioned observation for churn.

Looking at all the above example, it is evident that for any classification algorithm can classify well the outcome event but the difficult ask is, when the event is going to happen! In machine learning modelling there are some good ways to estimate the time to an event using hazard function modelling/ survival modelling to estimate the time to an event et el[19]. However, survival model suffers when the decision boundary becomes complex and results of the model starts deteriorating.

In this paper, we propose a novel approach to predict and event with some amount of lead time for business to take action.

This approach is inspired by reference frame et el[18]. This is commonly used in physics where a point of view/ coordinates /contexts, which is oriented to understand the variation or movement, subject to the frame of reference.

It can be explained with few examples: like a scenario of customer churn from a telecom industry. If someone has to just classify which customer may churn it's a straight forward classical modelling with any selected machine learning algorithm. However, the question is who are the customer like to churn by next month? This becomes not a straight forward classical modelling.

Another example let's assume a case of fault prediction model of network devices. Multiple variables can explain the impact on the DSLAM(Digital subscriber line access multiplexer) devices connects the network to the subscriber's premises. However if the modelling is done taking the data as it comes where at a time period algorithm is trained on the load, no of packet drop, signal to noise ratio (SNR) etc, algorithm would learn the behavior of time "T" as shown in the figure 1. Modelling on this data would generate coefficients or weights, which can classify the event, but lead-time is not guaranteed.

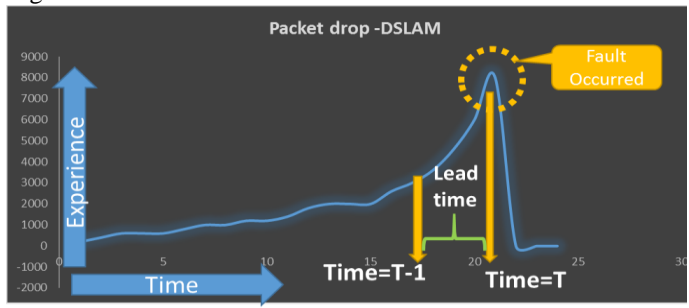

Fig. 1. Shows current classification model learns features based on point of impact as experience till the fault. Target in this paper to predict fault at T-1 period itself to gain some lead time to respond.

Our approach is to use the concepts of reference frame from physics and use it in the data preparation for the modelling so that in this process we can achieve some lead-time before the event occurs.

To achieve this we propose to use time "T-1" (some time ahead which is logical)as the reference frame for the modelling and accumulate the experience of the device/ built-up in each variable to predict the fault which can happen at time "T".

Critical part is to decide on the lead-time for the event. This is strictly based on the pattern of event for example in the case of fault prediction 1-3 hours of lead-time window has worked well for case studies we have worked on. However, for churn prediction of retail customer the lead-time can be 15-days to 30-days.

This is due to the fact that network fault has low/no relation with the values of the variable one month back compared to something which is more recent.

Basic idea of this approach is to treat the network devices as panel members and understanding it's behavior over time which cross-comparing with other devices would explain event occurrence.

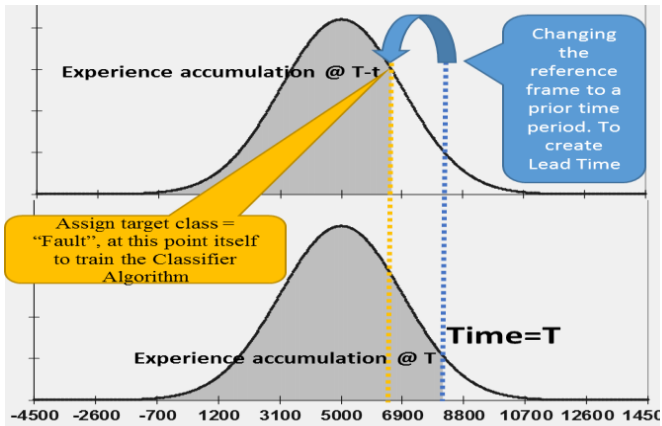

Fig. 2. Using the buildup (experience) of the object of study till prior to some lead time 't' and assigning target class based on time 'T' (fault occurred) to

Idea is to change the reference frame of the data from which the algorithm would learn. Let's assume we have x1,x2,x3….Xi variables explaining the occurrence of fault. Also let's assume the fault in one of the network device has happened at time "T" then if we have to create a lead time of "t" then we trace back in the data frame and find out what was the value of the variables during time "T-t" and assign the dependent variable level as "Fault" at that instance itself.

**Algorithm for Dynamic Reference Frame**

Require: Streaming data / Panel data of leading indicators or variables which is realtd to the event(fault/churn) $E_k$,

While event =1:
  for k in {x1,x2,x3....xk}
    value of xk = value of xk at time T-t
  End for
While event=0:
  for k in {x1,x2,x3....xk}
    value of xk = value of xk at time T
  End for

Illustrative example of the implementation:

Let's assume any supervised machine learning model to predict one dependent feature 'Y' based on 'm' independent feature(x). Algorithm provides key drivers estimate $\beta_k$ where $1 \leq k \leq m$ then the statistical model can be

### Raw data
#### Illustrative example of customer panel data

| Customer name | Month-Year | Outbound calls | No of complaints | No of Service Interruption | Fault resolution time | Promotion offered | Churn-Flag |
|---|---|---|---|---|---|---|---|
| Kumarjit | Jan-16 | 0 | 0 | 0 | 0 | 0 | 0 |
| Aasheesh | Jan-16 | 0 | 0 | 0 | 0 | 0 | 0 |
| Kumarjit | Feb-16 | 2 | 0 | 0 | 0 | 1 | 0 |
| Aasheesh | Feb-16 | 0 | 0 | 0 | 0 | 0 | 0 |
| Kumarjit | Mar-16 | 1 | 0 | 1 | 2 | 0 | 0 |
| Aasheesh | Mar-16 | 0 | 0 | 0 | 0 | 1 | 0 |
| Kumarjit | Apr-16 | 0 | 0 | 0 | 0 | 0 | 0 |
| Aasheesh | Apr-16 | 1 | 1 | 1 | 2 | 0 | 0 |
| Kumarjit | May-16 | 2 | 1 | 1 | 5 | 0 | 0 |
| Aasheesh | May-16 | 0 | 0 | 0 | 0 | 0 | 0 |
| Kumarjit | Jun-16 | 0 | 0 | 0 | 0 | 0 | 0 |
| Aasheesh | Jun-16 | 0 | 0 | 0 | 0 | 0 | 0 |
| Kumarjit | Jul-16 | 0 | 0 | 0 | 0 | 0 | 0 |
| Aasheesh | Jul-16 | 0 | 0 | 0 | 0 | 0 | 0 |
| Kumarjit | Aug-16 | 0 | 0 | 0 | 0 | 0 | 0 |
| Aasheesh | Aug-16 | 0 | 0 | 0 | 0 | 0 | 0 |
| Kumarjit | Sep-16 | 5 | 2 | 2 | 15 | 2 | 0 |
| Aasheesh | Sep-16 | 2 | 1 | 1 | 10 | 1 | 0 |
| Kumarjit | Oct-16 | 0 | 0 | 0 | 0 | 0 | 1 |
| Aasheesh | Oct-16 | 3 | 1 | 0 | 0 | 0 | 0 |
| Aasheesh | Nov-16 | 0 | 0 | 0 | 0 | 0 | 0 |
| Aasheesh | Dec-16 | 0 | 0 | 0 | 0 | 0 | 0 |
| Jitin | Dec-16 | 0 | 0 | 0 | 0 | 0 | 0 |
| Aasheesh | Jan-17 | 0 | 0 | 0 | 0 | 0 | 0 |
| Jitin | Jan-17 | 0 | 0 | 0 | 0 | 0 | 0 |
| Aasheesh | Feb-17 | 0 | 0 | 0 | 0 | 0 | 0 |
| Jitin | Feb-17 | 1 | 3 | 3 | 26 | 1 | 0 |
| Aasheesh | Mar-17 | 0 | 0 | 0 | 0 | 0 | 0 |
| Jitin | Mar-17 | 1 | 1 | 1 | 7 | 0 | 1 |
| Aasheesh | Apr-17 | 0 | 0 | 0 | 0 | 0 | 0 |
| Aasheesh | May-17 | 0 | 0 | 0 | 0 | 0 | 0 |
| Aasheesh | Jun-17 | 2 | 1 | 0 | 0 | 2 | 0 |
| Aasheesh | Jul-17 | 0 | 0 | 0 | 0 | 0 | 0 |
| Aasheesh | Aug-17 | 0 | 0 | 0 | 0 | 0 | 0 |
| Aasheesh | Sep-17 | 0 | 1 | 0 | 0 | 0 | 0 |
| Aasheesh | Oct-17 | 0 | 0 | 0 | 0 | 0 | 0 |
| Prabhu | Oct-17 | 2 | 1 | 0 | 0 | 0 | 0 |
| Aasheesh | Nov-17 | 0 | 0 | 0 | 0 | 0 | 0 |
| Prabhu | Nov-17 | 0 | 0 | 0 | 0 | 0 | 0 |
| Aasheesh | Dec-17 | 0 | 0 | 0 | 0 | 0 | 0 |
| Prabhu | Dec-17 | 0 | 0 | 0 | 0 | 0 | 0 |

With the above data Simple aggregation with the reference frame, when the customer has churned at time "T" accumulated experience of the customer would look like the table below:

| Customer name | Outbound calls _Total | Complaints _Total | Service Interruption _Total | Average Fault resolution time _total | Sum of Promotion offered _Total |
|---|---|---|---|---|---|
| Aasheesh | 8 | 5 | 2 | 6 | 4 |
| Jitin | 2 | 4 | 4 | 8.3 | 1 |
| Kumarjit | 12 | 4 | 5 | 4.4 | 3 |
| Prabhu | 2 | 1 | 0 | 0 | 0 |

However modelling on the above data does not guarantee for a lead-time in hand. To factor this lead-time component in to the model for classifying churn with the example data set, aggregation done based on one moth prior to the churn experience for the churners.

Eg: Kumarjit churned on Oct-16 so the data is aggregated until September-16 for Kumarjit; Similarly for Jitin who has churned on Mar-17, data is aggregated until Feb-17 only. Non-churner's data aggregation need not change.

| Customer name | Outbound calls _Total | Complaints _Total | Service Interruption _Total | Average Fault resolution time _total | Sum of Promotion offered _Total |
|---|---|---|---|---|---|
| Aasheesh | 8 | 5 | 2 | 6 | 4 |
| Jitin | 1 | 3 | 3 | 8.6 | 0 |
| Kumarjit | 10 | 3 | 4 | 5.5 | 3 |
| Prabhu | 2 | 1 | 0 | 0 | 0 |

This was an example of shifting the reference frame to one month prior to create the lead-time, however it would be problem specific and data scientist needs to take decision on the right amount of lead-time, which can be factored in, to the same. Ideally, more lead-time would deteriorate the accuracy of the model and hence there needs to be a tradeoff between amount of lead-time and the accuracy of the model.

## IV. CONCLUSION

In this paper we have integrated a concept from Physics to apply on regular statistical modelling. Dynamic reference frame can integrate best of both cross sectional and time series data variation.

Telecom vertical can use this to find exact cause of equipment failure for each of the equipment separately and provide support intelligence to engineers and reduce cycle time of repair.

Retail sector can use this strategy to proactively identify customer who are likely to churn in next month and initiate communication with the customers before the decision is taken.

REFERENCES

[1] Jingjiao Zhang, Jiaqing Fu , Chunhong Zhang , Xin Ke , Zheng Hu, "Not Too Late to Identify Potential Churners: Early Churn Prediction in Telecommunication Industry" IEEE , 16 March 2017, INSPEC Accession Number: 16757983.
[2] Clifton Phua, Hong Cao, João Bártolo Gomes, Minh Nhut Nguyen "Predicting Near-Future Churners and Win-Backs in the Telecommunications Industry" Data Analytics Department, Institute for Infocomm Research, 1 Fusionopolis Way, Connexis, Singapore 38632.
[3] V. Umayaparvathi1, K. Iyakutti, "A Survey on Customer Churn Prediction in Telecom Industry: Datasets, Methods and Metrics" Apr-2016 International Research Journal of Engineering and Technology (IRJET).
[4] Amoo A. O, Akinyemi B. O, Awoyelu I. O, Adagunodo E. R " Modeling & Simulation of a Predictive Customer Churn Model for Telecommunication Industry" Journal of Emerging Trends in Computing and Information Sciences Nov-2015.
[5] Imran Khan, Imran Usman, Tariq Usman, Ghani Ur Rehman, and Ateeq Ur Rehman "Intelligent Churn prediction for Telecommunication Industry" International Journal of Innovation and Applied Studies, Sep 2013.
[6] Ali Tamaddoni Jahromi - "Predicting Customer churn in telecommunication service" Luela University of Technology , [url: https://www.researchgate.net/profile/Ali_Tamaddoni/publication/267196933_Predicting_Customer_Churn_in_Telecommunications_Service_Providers/links/5555860c08ae6943a871c662/Predicting-Customer-Churn-in-Telecommunications-Service-Providers.pdf].
[7] Pradeep B, Sushmitha Vishwanath Rao and Swati M Puranik "Analysis of Customer Churn prediction in Logistic Industry using Machine Learning" , International Journal of Scientific and Research Publications, Volume 7, Issue 11, November 2017.
[8] Muhammad Raza Khan, Joshua Manoj, Anikate Singh, Joshua Blumenstock – "Behavioral Modeling for Churn Prediction" , October 18, 2016. [URL: https://pdfs.semanticscholar.org/339f/de557d08aa9612ce7f372fb312d78dce6a27.pdf]
[9] Anonymous , "Sample Classification Predictions" [url: http://wiki.eigenvector.com/index.php?title=Sample_Classification_Predictions]
[10] Anonymous, "Prediction Lag" [url: https://wiki.seg.org/wiki/Prediction_lag]
[11] Shin-Yuan Hung a, David C. Yen b, Hsiu-Yu Wang "Applying data mining to telecom churn management"
[12] Guangli Nie , Wei Rowe , Lingling Zhang ,Yingjie Tian , Yong Shi, " Credit card churn forecasting by logistic regression and decision tree" Science Direct 2011
[13] SCOTT A. NESLIN, SUNIL GUPTA, WAGNER KAMAKURA, JUNXIANG LU, and CHARLOTTE H. MASON "Defection Detection: Measuring and Understanding the Predictive Accuracy of Customer Churn Models"
[14] Wouter Verbeke , David Martens , Christophe Mues , Bart Baesens , "Building comprehensible customer churn prediction models with advanced rule induction techniques" Science Direct ,2011
[15] Hamidreza Zareipour, Dongliang Huang, William Rosehart, "Wind Power Ramp Events Classification and Forecasting" IEEE 2011
[16] JE Boylan, AA Syntetos and GC Karakostas "Classification for forecasting and stock control:a case study" Journal of the Operational Research Society (2008) 59, 473 –48
[17] Nikolas Roman Herbst, Nikolaus Huber, Samuel Kounev, and Erich Amrehn "Self-Adaptive Workload Classification and Forecasting for Proactive Resource Provisioning" 2014 [url: https://pdfs.semanticscholar.org/9b55/aea33dfe3a4e9a661f6b1f34fcaa8abba6c7.pdf]
[18] "Theory of reference frame" [url: http://factmyth.com/reference-frames-examples/]
[19] "Survival Models", [url: http://data.princeton.edu/wws509/notes/c7.pdf]